%% file: main.tex
\documentclass[11pt]{article}

\usepackage{acl}

\usepackage{array}

\usepackage{times}
\usepackage{latexsym}

\usepackage[T1]{fontenc}

\usepackage[utf8]{inputenc}

\usepackage{microtype}

\usepackage{inconsolata}

\usepackage{svg}
\usepackage{amsmath}
\usepackage{tcolorbox}
\usepackage{booktabs}            
\usepackage{tabularx}            
\usepackage{siunitx}       
\usepackage{subcaption}
\usepackage{graphicx}     
\usepackage{lipsum}
\usepackage{multirow}
\usepackage{array}
\usepackage{xcolor}
\usepackage{soul}
\usepackage{makecell}
\sethlcolor{green!15}

\tcbuselibrary{breakable}

\tcbset{
    scenariobox/.style={
        colback=green!5,
        colframe=green!60!black,
        boxrule=0.7pt,
        arc=2mm,
        left=6pt,
        right=6pt,
        top=6pt,
        bottom=6pt
    }
}

\title{Emotion Entanglement and Bayesian Inference for Multi-Dimensional Emotion Understanding}

\author{
\textbf{Hemanth Kotaprolu}$^{\dagger}$,
\textbf{Kishan Maharaj}$^{\mathsection\dagger}$
\textbf{Raey Zhao}$^{\ddagger}$, \\
\textbf{Abhijit Mishra}$^{\ddagger}$,
\textbf{Pushpak Bhattacharyya}$^{\dagger}$ \\
$^{\dagger}$Indian Institute of Technology Bombay, Mumbai, India \\
$^{\ddagger}$University of Texas at Austin, Texas, United States \\
$^{\mathsection}$IBM Research \\
\texttt{23m2164@iitb.ac.in, kishanmaharaj@ibm.com, zhao993059@utexas.edu} \\
\texttt{abhijitmishra@utexas.edu, pb@cse.iitb.ac.in}
}

\begin{document}

\maketitle

\begin{abstract}
Understanding emotions in natural language is inherently a multi-dimensional reasoning problem, where multiple affective signals interact through context, interpersonal relations, and situational cues. However, most existing emotion understanding benchmarks rely on short texts and predefined emotion labels, reducing this process to independent label prediction and ignoring the structured dependencies among emotions. To address this limitation, we introduce \textbf{Emo}tional \textbf{Scen}arios (\textbf{EmoScene}), a theory-grounded benchmark of 4,731 context-rich scenarios annotated with an \textit{8-dimensional emotion vector derived from Plutchik's basic emotions}. Motivated by the observation that emotions rarely occur independently, we further propose an \textit{entanglement-aware Bayesian inference} framework that incorporates emotion co-occurrence statistics to perform joint posterior inference over the emotion vector. This lightweight post-processing does not require any parameter updates and improves the structural consistency of predictions, and yields overall gains of 2.24\% Lexical Accuracy without any additional cost. EmoScene therefore provides a challenging benchmark for studying \textit{multi-dimensional emotion understanding} and the limitations of current language models.

\end{abstract}

\input{01_introduction}
\input{02_related_work}

\input{03_emoscene_dataset}
\input{04_methodology}

\input{05_experimental_setup}

\input{06_results_and_analysis}

\input{07_summary_and_conclusion}
\input{08_limitations}

\bibliography{main}

\appendix
\input{09_appendix}

\end{document}

%% file: 01_introduction.tex
\section{Introduction}
\label{sec:intro}
Emotional intelligence (EI) refers to the ability to recognize, understand, and regulate emotions in oneself and others \cite{Mayer2004-dt}. Emotions play a fundamental role in human cognition, influencing decision making, social interaction, and behavioural responses to environmental stimuli \cite{Hieida2018DeepEA}. Because emotional signals are often implicit and context-dependent, understanding them requires interpreting subtle linguistic cues, situational context, and interpersonal relationships. In NLP, this ability is typically evaluated through emotion classification tasks that aim to measure a system's capacity for \textit{Emotion Understanding} (EU), namely the interpretation of emotions expressed within a textual context \cite{sabour-etal-2024-emobench, Buechel2017EmoBankST}. Despite significant progress in affective computing, modelling emotions remains challenging because emotional expressions are subjective, context-sensitive, and often intertwined with social dynamics \cite{Izard2002-ph, Perloff1997-za}. As the existing NLP systems become increasingly capable of producing fluent and contextually appropriate responses, it becomes more important to know if LLMs genuinely understand emotions, or do they merely reproduce linguistic patterns associated with emotional language \cite{sabour-etal-2024-emobench}?

\begin{figure*}
    \centering
    \includegraphics[scale=0.26]{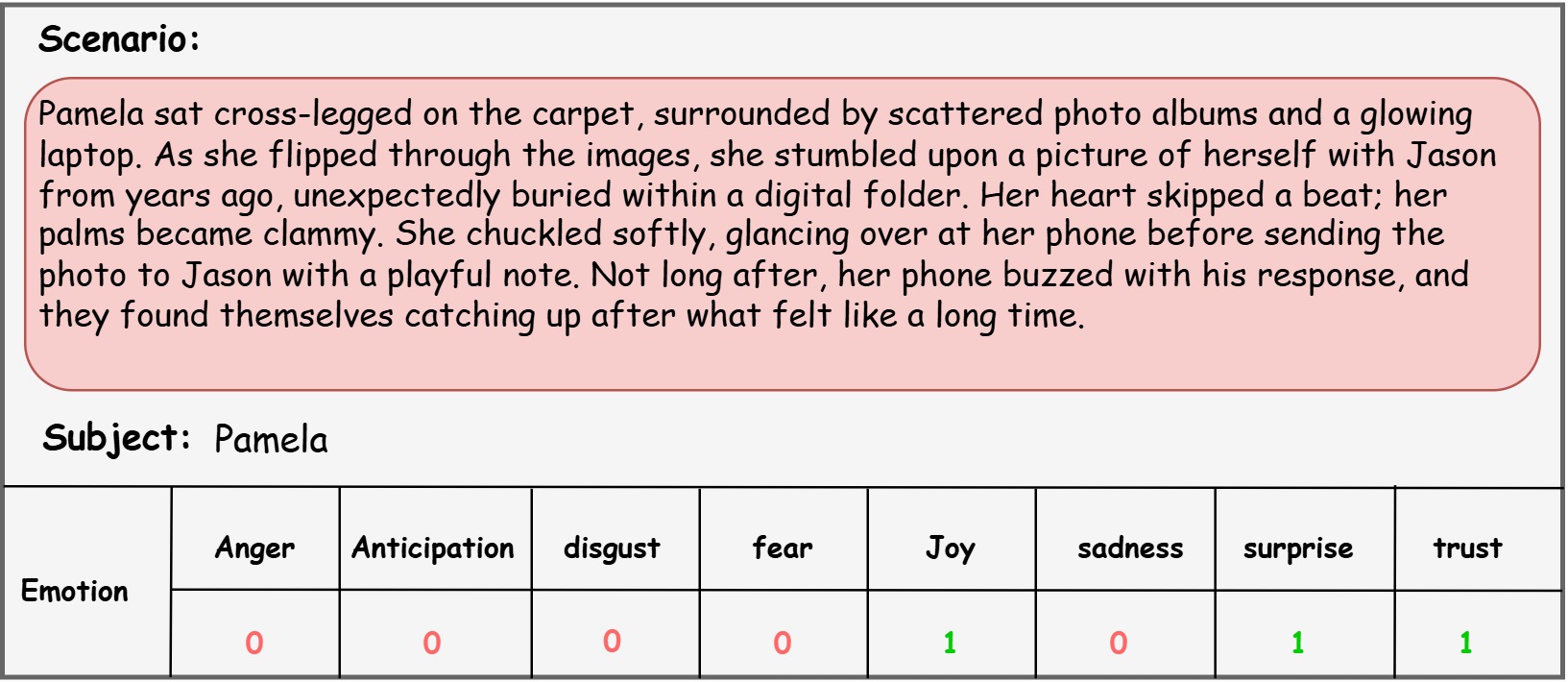}
  \caption{An example scenario from the EmoScene Dataset. The ground truth labels Joy, Surprise, and Trust must be inferred from behavioral and situational cues. The unexpected discovery ("stumbled upon", "heart skipped a beat") signals Surprise, the soft chuckle and playful outreach signal Joy, and the willingness to reconnect with Jason after a long gap signals Trust. Attending only to surface cues ("heart skipped a beat", "clammy palms") alone risks predicting Fear; the surrounding behavioral and relational context is what disambiguates the true affect.}
  \label{fig: example_scenario}
  \end{figure*}
  

Datasets such as SemEval-2018 Task 1: Affect in Tweets \cite{mohammad-etal-2018-semeval}, GoEmotions \cite{demszky-etal-2020-goemotions}, ISEAR \cite{troiano-etal-2019-crowdsourcing}, and EmotionLines \cite{chen2018emotionlinesemotioncorpusmultiparty} have enabled large-scale evaluation of emotion classification systems. However, many of these datasets suffer from key limitations. First, inputs are often short sentences or social media posts that lack sufficient context to disambiguate subtle emotional cues. Second, they rely on predefined emotion labels that may not align with established psychological theories of emotion. Third, most evaluation protocols treat emotions as independent categories, ignoring the fact that emotional states frequently co-occur and influence one another. Recent work such as \cite{sabour-etal-2024-emobench} attempts to make emotion understanding and application tasks more implicit; however, it is limited to single-label settings and does not account for the co-occurrence of multiple emotions. Additionally, the proposed dataset is relatively small in scale, comprising only 200 instances.    

In this work, we introduce Emotional Scenarios (EmoScene), a large-scale dataset comprising 4.7K instances, designed to evaluate emotion understanding in complex, context-rich settings while explicitly accounting for the co-occurrence of multiple emotions. Each instance consists of a subject paired with a detailed scenario that captures environmental cues, physiological responses, and interpersonal interactions. We ground the annotation scheme in Plutchik's theory of emotions \cite{plutchik1980general} and represent each scenario using an \textbf{8-dimensional binary emotion vector corresponding to Plutchik's basic emotions} (ref. example from Figure \ref{fig: example_scenario}. This formulation enables the modelling of complex emotional states as combinations of fundamental emotional dimensions, which is essential for multi-label settings. In contrast to \cite{sabour-etal-2024-emobench}, which uses Plutchik's theory for developing a taxonomy that collapses multiple emotions into a single label, our approach preserves the compositional structure of emotional experiences. Furthermore, recognising that emotions rarely occur in isolation, we formulate emotion prediction as a structured inference problem. To this end, we propose an entanglement-aware Bayesian inference framework that jointly infers the emotion vector's posterior. This lightweight approach ($\sim {6*10^{-4}}$ sec overhead per scenario) requires no parameter updates and explicitly captures dependencies among emotions, thereby addressing the independence assumptions commonly made in standard multi-label classification. As a post-processing step, it improves prediction consistency and yields an overall gain of 2.24\% in lexical accuracy (CoT.


\textbf{Our contributions are summarized as follows:}

\begin{enumerate}
    \item \textbf{EmoScene Benchmark.} A context-rich dataset of \textbf{4,731 scenarios} for evaluating emotion understanding in natural language grounded on Plutchik's basic emotions, where emotion prediction is formulated as an \textbf{8-dimensional multi-label task} 
    
    \item \textbf{Emotion entanglement modelling} A \textbf{Bayesian inference framework} that incorporates statistical dependencies between emotions for joint prediction.
    
    \item \textbf{LLM evaluation.} A systematic zero-shot study of \textbf{seven instruction-tuned LLMs} on multi-dimensional emotion understanding.
\end{enumerate}

The code and data have been released for academic use at: \href{https://anonymous.4open.science/r/EmoScene-E6F8/}{EmoScene-Codes}.

%% file: 02_related_work.tex
\begin{figure*}
    \centering
    \includegraphics[scale=0.18]{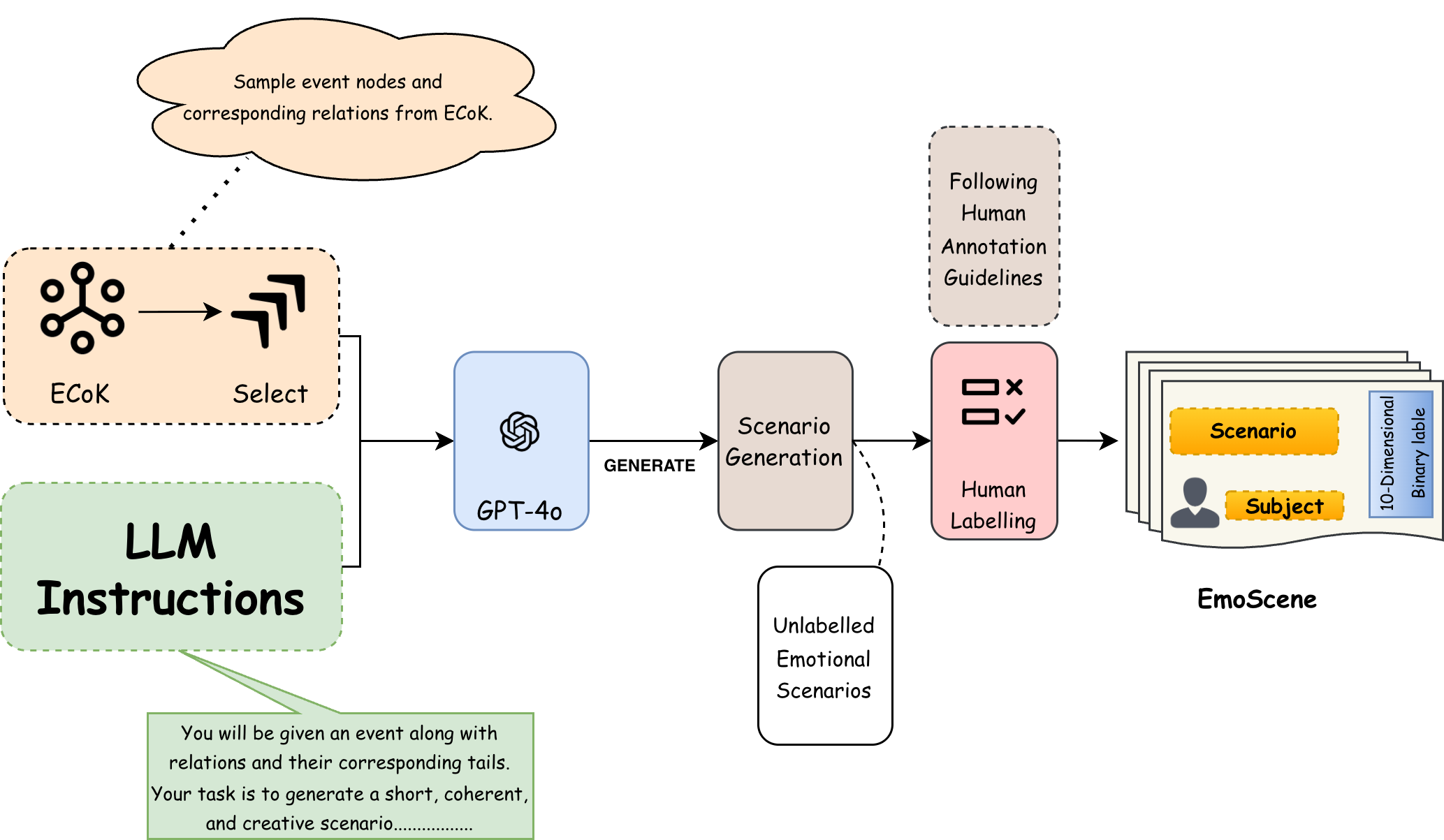}
  \caption{Overview of our steps to construct EmoScene dataset}
  \label{fig:schematic}
\end{figure*}

\section{Related Work}
\label{sec:relwork}

\paragraph{Emotion theories and psychological grounding.}
Emotions have long been studied in psychology as adaptive mechanisms that influence cognition, behavior, and social interaction \cite{damasio_emotion_1998, Gross2011-ht}. Emotional intelligence refers to the capacity to recognize, interpret, and manage emotional information in oneself and others \cite{Mayer2004-dt}. A large body of work has explored how emotional experiences are structured and expressed in human cognition. For example, emotional granularity describes the ability to distinguish subtle affective states and has been linked to improved emotional regulation and social functioning \cite{Hoemann2021-xs}. Psychological theories such as Plutchik's psychoevolutionary theory further suggest that complex emotional states arise from combinations of a small set of fundamental emotions \cite{Plutchik1982-pc, Plutchik1980-en}; this model has been a commonly used grounding in recent work. These theoretical perspectives highlight that emotions are inherently structured and often co-occur rather than appearing in isolation.

\paragraph{Emotion Understanding in NLP.}
Emotion analysis has evolved from lexicon-based classification \cite{Mohammad13} to context-dependent, perspective-aware, and multimodal recognition \cite{Fu2023-me, yeo-jaidka-2023-peace, yeo2025understanding, liu2024speak, chandraumakantham2024multimodal}. To capture interconnected affective signals, recent frameworks emphasize multi-label settings and structured relationships, such as Plutchik-grounded models \cite{belay-etal-2025-evaluating, xue-etal-2025-jnlp, chang-BEAM}. Concurrently, the rise of large language models (LLMs) has shifted research toward evaluating their broader capacity for emotional intelligence, including affect-aware reasoning and empathy \cite{sabour-etal-2024-emobench, sorin2024large}. While studies indicate LLMs can internally represent and process ambiguous emotional cues \cite{hong2025ambiguity-aware, reichman2026emotion}, their performance remains highly sensitive to contextual nuances and evaluation design, highlighting the need for more structured benchmarks, especially those that highlight the complexities arising from co-occurrences of emotions.

\paragraph{Emotion datasets and benchmarks.}
Early benchmarks focused on identifying emotions in short texts like tweets \cite{strapparava-mihalcea-2007-semeval, mohammad-etal-2018-semeval}. Subsequent datasets, such as GoEmotions \cite{demszky-etal-2020-goemotions}, EmotionLines \cite{chen2018emotionlinesemotioncorpusmultiparty}, and ISEAR \cite{troiano-etal-2019-crowdsourcing, Scherer1990International}, expanded taxonomies into conversational contexts. While some work has explored long-form multi-class analysis \cite{liu-etal-2019-dens} and evolving emotions \cite{wemmer-etal-2024-emoprogress}, the majority of datasets still rely on short, decontextualized inputs and rigid, predefined categories that struggle to capture complex emotional states. Furthermore, varied annotation practices—ranging from crowdsourcing to expert labeling—often compromise label consistency and reliability.

\paragraph{Limitations of existing benchmarks and our approach.}
Existing benchmarks often treat emotions independently, ignoring the structured dependencies of human affect and obscuring whether models genuinely reason about emotions. Further, \textit{to the best of our knowledge, there is no emotion understanding benchmark that captures bodily symptoms, action tendencies, while being multilabel}. To address these limitations, we introduce \textbf{EmoScene}, a benchmark of context-rich scenarios annotated with an 8-dimensional emotion vector grounded in Plutchik's theory. We explicitly model these dependencies using a Bayesian inference framework that incorporates empirical co-occurrence patterns, thereby framing emotion understanding as a structured multi-label inference problem.

%% file: 03_emoscene_dataset.tex
\section{EmoScene Dataset}
\label{sec:03_emoscene_dataset}
The EmoScene benchmark consists of $4,731$ scenarios designed to evaluate multi-dimensional emotion understanding. Each instance consists of a subject with a detailed scenario describing environmental cues, physiological responses, and interpersonal interactions. Following Plutchik’s psychoevolutionary theory \cite{Plutchik1980-en}, we annotate each scenario with an \textbf{8-dimensional binary emotion vector} where each dimension corresponds to Plutchik's basic emotions and allows co-occurrence of multiple labels. By preserving the compositional structure of emotional experiences, EmoScene enables the evaluation of complex emotional states as multi-label prediction tasks.


The dataset was constructed in two stages: \textbf{scenario generation} and \textbf{human annotation} (Figure~\ref{fig:schematic}). For scenario generation, we leverage the \textbf{Emotional Commonsense Knowledge Graph (ECoK)} \cite{wang-etal-2024-ecok}, which consists of \texttt{<head, relation, tail>} triples. The head entities describe emotional events, while seven relation types (e.g., \textit{causes, bodily symptoms, action tendencies}) capture different facets of the emotional experience. We leverage ECoK relations, specifically feelings (for subjective experiences), bodily symptoms (for physiological responses) and facial expressions (for the behavioural manifestation), which is crucial to appropriately describe subtle cues required for accurate emotion understanding. Specifically, this method of scenario generation ensures that its structured schema embeds situational, physiological, and interpersonal cues, yielding richer and more realistic context than short-text or social media datasets.

Using these triples as conceptual seeds, we prompt GPT-4o to generate the context-rich scenarios (the full generation prompt is provided in Appendix~\ref{app:emoscene-generation-prompt}). A comprehensive analysis of the dataset—including dataset statistics, topic diversity, and label distributions—is detailed in Appendices~\ref{app:dataset_statistics} and \ref{app:dataset_analysis}.

\textbf{Annotation and inter-annotator agreement}
Three expert annotators (English instructors with prior annotation experience) independently labelled every scenario by answering eight yes/no questios, one per emotion. Final Labels were assigned by strict majority vote. Before annotation, a training session has been conducted individuall to all annotators which involves explaining the guidelines and providing definitions of various emotions. Inter-annotator agreement (Fleiss' $\kappa$) varies by emotion, yielding a mean of $0.43$. This is consistent with prior popular subjective-emotion benchmarks such as GoEmotions \cite{demszky-etal-2020-goemotions} and falls in the moderate range described by \citet{mchugh2012interrater}. Notably, Trust is the most prevalent emotion in EmoScene ($48.1\%$), yet it proves the most difficult to label consistently and is frequently missed by LLMs (see \S\ref{sec:error_analysis}). Annotators are compensated accordingly.



%% file: 04_methodology.tex
\section{Methodology}
\label{04_methodology}

Traditional multi-label frameworks typically model emotions as independent Bernoulli variables, failing to capture the structured interdependencies nature of human affect, where emotions are rarely independent and often co-occur in structured, interdepended ways. To address this limitations, we introduce Emotion Entanglement, a perspective that treats emotional states as a coupled system rather than disjoint labels. Building on this perspective, we propose an Entanglement-Aware Bayesian Inference framework that explicitly models the joint distribution over emotions, incorporating co-occurrence dependencies into the inference process to produce more coherent, context-sensitive predictions without requiring expensive parameter updates.



\begin{table*}[t]
\centering
\footnotesize
\setlength{\tabcolsep}{5pt}
\begin{tabular}{l cc cc cc cc}
\toprule
& \multicolumn{2}{c}{\textbf{Lexical Accuracy}}
& \multicolumn{2}{c}{\textbf{Vector Accuracy}}
& \multicolumn{2}{c}{\textbf{Hamming Loss}}
& \multicolumn{2}{c}{\textbf{F1-Macro}} \\
\cmidrule(lr){2-3}\cmidrule(lr){4-5}\cmidrule(lr){6-7}\cmidrule(lr){8-9}
\textbf{Model}
& \textbf{Baseline} & \textbf{BI}
& \textbf{Baseline} & \textbf{BI}
& \textbf{Baseline} & \textbf{BI}
& \textbf{Baseline} & \textbf{BI} \\
\midrule

\multicolumn{9}{c}{\textbf{Chain-of-Thought (CoT) Prompting}} \\
\midrule
Llama-3.1-8B & 83.87 & 83.88 & 24.54 & 24.50 & 16.13 & 16.12 & 47.2 & 47.2 \\
Llama-3.2-3B & 79.76 & 79.76 & 17.40 & 17.40 & 20.24 & 20.24 & 45.1 & 45.1 \\
Llama-3.1-70B & 81.93 & 83.27 & 22.15 & 23.95 & 18.07 & 16.73 & 50.6 & 51.7 \\
Mistral-Nemo & 83.48 & 83.50 & 22.24 & 22.26 & 16.52 & 16.50 & 43.1 & 43.2 \\
Qwen2.5-7B   & 75.95 & 82.22 & 14.06 & 21.48 & 24.05 & 17.78 & 42.8 & 47.9 \\
Gemma-2-2b   & 70.93 & 78.79 & 10.27 & 15.96 & 29.07 & 21.21 & 43.7 & 47.4 \\
Gemma-2-9b   & 83.37 & 83.48 & 24.88 & 25.07 & 16.63 & 16.52 & 50.0 & 50.1 \\

\midrule
\textit{Overall} & 79.89 & 82.13 {\scriptsize (\textbf{+2.24})} & 19.36 & 21.51 {\scriptsize (\textbf{+2.15})} & 20.11 & 17.87 {\scriptsize (\textbf{$-$2.24})} & 46.07 & 47.51 {\scriptsize (\textbf{+1.44})} \\

\midrule
\multicolumn{9}{c}{\textbf{Zero-Shot Prompting}} \\
\midrule
Llama-3.1-8B & 77.92 & 82.04 & 14.61 & 20.10 & 22.08 & 17.96 & 50.5 & 53.0 \\
Llama-3.2-3B & 83.43 & 83.53 & 24.12 & 24.33 & 16.57 & 16.47 & 48.3 & 48.4 \\
Llama-3.1-70B & 75.57 & 80.91 & 10.95 & 17.42 & 24.43 & 19.09 & 50.2 & 53.7 \\
Mistral-Nemo & 83.54 & 83.54 & 21.69 & 21.69 & 16.46 & 16.46 & 44.0 & 44.0 \\
Qwen2.5-7B   & 83.32 & 83.36 & 22.09 & 22.13 & 16.68 & 16.64 & 44.7 & 44.8 \\
Gemma-2-2b   & 77.21 & 77.57 & 13.89 & 14.23 & 22.79 & 22.43 & 42.8 & 43.1 \\
Gemma-2-9b   & 81.67 & 82.68 & 21.71 & 22.87 & 18.33 & 17.32 & 50.6 & 51.1 \\
\midrule
\textit{Overall} & 80.38 & 81.95 {\scriptsize (+1.57)} & 18.43 & 20.39 {\scriptsize (+1.96)} & 19.62 & 18.05 {\scriptsize ($-$1.57)} & 47.3 & 48.3 {\scriptsize (+1.00)} \\
\bottomrule
\end{tabular}
\caption{Model Performance Comparison on EmoScene. We compare Baseline outputs against our Entanglement-Aware Bayesian Inference (BI) post-processing step across both Chain-of-Thought and Zero-Shot settings.}
\label{tab:complete_results_table}
\end{table*}

\subsection{Modeling Emotion Entanglement}
We define \textbf{Emotion Entanglement} as the phenomenon where the presence of one emotion in a text significantly alters the likelihood of another co-occurring. Formally, we hypothesize that for an input sentence $x$ and an emotion vector $E = (E_1, \dots, E_{8})$, the joint distribution does not factorize:

\begin{equation}
P(E|x) \neq \prod_{i=1}^{8} P(E_i|x)
\end{equation}

To quantify this, we analyse the structural non-independence of emotions using information-theoretic measures. Specifically, we compute the Mutual Information $I(E_i; E_j)$ between emotion pairs to establish that emotional expression is structurally dependent. For instance, in the EmoScene dataset, the probability of \textit{Anger} is significantly higher given the presence of \textit{Sadness} than its marginal probability, i.e., $P(\text{Anger}=1 | \text{Sadness}=1) \gg P(\text{Anger}=1)$.

\subsection{Entanglement-Aware Bayesian Inference}
To correct the independence assumption made by vanilla LLMs, we propose a post-processing step based on Bayesian inference. We treat the LLM output as a noisy likelihood and combine it with a data-derived prior that captures label entanglement.

Let $x$ be the input text and $E = (E_1, \dots, E_{8})$ be the binary emotion vector where $E_i \in \{0, 1\}$. Our goal is to find the Maximum A Posteriori (MAP) estimate $\hat{E}$:

\begin{equation}
\hat{E} = \arg \max_{E} P(E|x) \propto \arg \max_{E} P(x|E) \cdot P(E)
\end{equation}

\subsubsection{Estimating the Entanglement Prior}
We model the label-only prior $P(E)$ using an \textbf{Ising Model} (a type of Markov Random Field) to find the least-committed distribution that matches observed marginals and pairwise correlations in the training data. The prior is defined as:

\begin{equation}
P(E) = \frac{1}{Z} \exp \left( \sum_{i} \theta_i E_i + \sum_{i<j} \theta_{ij} E_i E_j \right)
\end{equation}

where $\theta_i$ represents the bias for individual emotions and $\theta_{ij}$ represents the coupling strength between emotion pairs. These parameters are estimated from the training set via maximum-entropy solutions:

\begin{equation}
\begin{aligned}
\theta_i &= \log \frac{\hat{P}(E_i=1)}{1-\hat{P}(E_i=1)}, \\
\theta_{ij} &= \log \frac{\hat{P}(E_i=1, E_j=1)}
{\hat{P}(E_i=1)\hat{P}(E_j=1)}
\end{aligned}
\end{equation}

Critically, these learned couplings $\theta_{ij}$ empirically recover the dyadic structure of Plutchik's wheel, such as strong mutual exclusion of bipolar opposites (see Appendix \ref{app:plutchik_alignment} for more details)

\subsubsection{Likelihood and MAP Inference}
The likelihood $P(x|E)$ is derived from the LLM's log-probabilities for each emotion category. Assuming a Bernoulli likelihood for the text given an emotion configuration:

{\small
\begin{equation}
\log P(x|E) = \sum_{i} \left[ E_i \log p_i^{(1)} + (1-E_i) \log p_i^{(0)} \right]
\end{equation}
}

Substituting the prior and likelihood back into the MAP objective, our final a-posteriori log-objective becomes:

\begin{equation}
\begin{aligned}
\log P(E|x) = {} & \sum_{i} \left[ E_i \log \frac{p_i^{(1)}}{p_i^{(0)}} \right] \\
& + \alpha \left[ \sum_{i} \theta_i E_i 
+ \sum_{i<j} \theta_{ij} E_i E_j \right]
\end{aligned}
\end{equation}

Since we utilize a fixed space of 8 labels, the exact argmax is found by evaluating all $2^{8}$ combinations, ensuring a parameter-efficient post-processing step that explicitly models emotion co-occurrence fidelity.


%% file: 05_experimental_setup.tex
\section{Experimental Setup}
\label{05_exp_setup}
In this section, we describe the models and hyperparameters that we have used to run the experiments. Details about confidence extraction is provided in the Appendix~\ref{app:extended_exp_setup}

\paragraph{Models Details.} We evaluate seven open-source instruction-tuned LLMs from HuggingFace—\texttt{Mistral-Nemo-2407-Instruct}, \texttt{Llama-3.2-3B-Instruct}, \texttt{Gemma-2-2B-it}, \texttt{Qwen2.5-7B-Instruct}, \texttt{Gemma-2-9B-it}, and \texttt{Llama-3.1-8B-Instruct}, \texttt{Llama-3.1-70B-Instruct}—run locally on $4\times$ NVIDIA A100 80GB GPUs. 

\paragraph{Evaluation Metrics.} We evaluate performance using \textbf{Lexical Accuracy (LA)} (fraction of correctly predicted emotion cells), \textbf{Vector Accuracy (VA)} (exact-match fraction of 8-dimensional vectors), and \textbf{Hamming Loss (HL)} ($1 - \text{LA}$). Our primary metric is \textbf{Macro F1}, which averages the F1 scores across all eight emotions equally, ensuring performance on rare emotions (e.g., Disgust, Surprise) is weighted equally to common ones \cite{Mohammad13}.

\paragraph{Confidence Extraction and Bayesian Inference.} The Bayesian Inference (BI) step relies on a Bernoulli likelihood $P(x|E)$, populated by softmaxing the model's \texttt{yes}/\texttt{no} logits at the answer position. The Ising prior ($\theta_i, \theta_{ij}$) is derived from the full EmoScene label matrix. We perform exact inference by enumerating all $2^8 = 256$ label configurations and sweep the hyperparameter $\alpha \in \{0, 0.1, 0.25, 0.5, 0.75, 1.0, 2.0, 5.0\}$ to find the per-model F1-optimal value (see \S\ref{sec:alpha_sensitivity}).

%% file: 06_results_and_analysis.tex
\section{Results \& Analysis}
\label{06_results_and_analysis}

\begin{table}[t]
\centering
\footnotesize
\setlength{\tabcolsep}{4pt} 
\begin{tabular}{l c c c c}
\toprule
\textbf{Model} & $\alpha$ & \makecell{\textbf{Baseline} \\ \textbf{F1}} & \makecell{\textbf{Linear} \\ \textbf{F1}} & \makecell{\textbf{Full-prior} \\ \textbf{F1}} \\
\midrule
Llama-3.1-8B & 0.10 & 47.15 & 47.15 & 47.16 \\
Llama-3.2-3B & 0.00 & 45.13 & 45.13 & 45.13 \\
Llama-3.1-70B & 0.50 & 50.60 & 51.50 & 51.75 \\
Mistral-Nemo & 0.25 & 43.10 & 43.15 & 43.17 \\
Qwen2.5-7B   & 0.75 & 42.81 & 46.15 & 47.89 \\
Gemma-2-2B   & 0.75 & 43.68 & 46.17 & 47.36 \\
Gemma-2-9B   & 0.50 & 50.02 & 49.98 & 50.07 \\
\bottomrule
\end{tabular}
\caption{Effect of ablating the pairwise term $\theta_{ij}$ at each model's F1-optimal $\alpha$. }
\label{tab:linear_only_ablation}
\end{table}

The main results in Table~\ref{tab:complete_results_table} contain the evaluation results for both prompting strategies: CoT \& ZS. For both prompts we compare the LLM's baseline prediction against our Bayesian Inference (BI) post-processing. Evaluation prompt can be found in Figure~\ref{fig:evaluation_prompt} \& Figure~\ref{fig:zeroshot_evaluation_prompt}. More details on Prompting Strategies, Significance of $\alpha$ and confidence calibration are provided in the Appendix~\ref{app:extended_results}


Across the six open-source models, Bayesian Inference (BI) delivers its largest gains on the baselines that most over-predict rare emotions: \textbf{Qwen2.5-7B} ($+5.1$ F1, $+6.27$ LA) and \textbf{Gemma-2-2B} ($+3.7$ F1, $+7.86$ LA). BI's marginal prior $\theta_i$ pulls these over-predictions toward the empirical zero-rate, recovering F1 without retraining. For the four better-calibrated baselines (baseline LA $> 79\%$), gains are smaller as there is less over-prediction to correct; on Llama-3.2-3B, the optimal $\alpha$ is 0 and BI has no effect. Llama-3.1-70B achieves the highest baseline Macro F1 (\textbf{50.6}) narrowly beating the best open-source model, Gemma-2-9B which has 50.0 MacroF1 confirming EmoScene's difficulty.  

Despite high lexical scores, Vector Accuracy remains modest ($15\text{--}25\%$). Because BI performs joint MAP inference over the full emotion vector $\mathbf{E}$, it enforces an entanglement prior via an MRF/Ising graph that models pairwise correlations. This corrects the LLM independence assumption by re-weighting predictions to match real co-occurrence patterns. BI consistently improves Hamming Loss and Macro F1, demonstrating BI's effectiveness for robust, scene-aware emotion classification.

\paragraph{Where BI does and does not help?}
For baselines that are already well-calibrated to the emotion marginals (Llama-3.1-8B, Gemma-2-9B), per-emotion precision (refer Appendix~\ref{sec:per_emotion}) is already high and BI has little room to operate consistent with the small overall gains in Table~\ref{tab:complete_results_table}. For baselines whose lexical-accuracy is dragged down by over-prediction on rare emotions (Qwen2.5-7B, Gemma-2-2B), BI recovers $+5$ to $+8$ LA points specifically by pruning these false positives. 

\begin{tcolorbox}[scenariobox,title=\textbf{Scenario [Anticipation, Sadness]}]
Sonia scrolled through her Instagram feed, her eyes narrowing as she paused on Frank's latest post about his trip to Bali. Her stomach felt unsettled, and she noticed her muscles tensing as she read about his adventures. She couldn't help but compare her own routine life to his glamorous lifestyle. Trying to shake off the feeling, she decided to focus on her own upcoming weekend plans and the things she has achieved recently.
\end{tcolorbox}


In the above scenario, the baseline incorrectly predicts Anger along with Anticipation and Sadness. In general, Sadness is a low-arousal negative emotion, whereas Anger is a high-arousal negative emotion. The description of her stomach being "unsettled" and muscles "tensing" injects a high level of physical arousal into the text. The baseline model struggled to map this high-arousal discomfort to Sadness, defaulting instead to Anger, which is the most common high-arousal negative emotion in text classification. Our formulation removes this spurious label of Anger driven by the learned prior, where $\theta_{ij}$ favours the co-occurrence of Sadness and Anticipation but penalises unlikely associations with Anger. 

\paragraph{Efficiency Analysis} To evaluate BI’s inference overhead, we measure the additional latency using the largest model, \texttt{Llama-3.1-70B}. BI adds only ${6 \times 10^{-4}}$ seconds per scenario on average, indicating negligible inference overhead.

\subsection{Error Analysis}
\label{sec:error_analysis}

To better understand where LLMs fail on EmoScene, we systematically rank all candidate error patterns by their frequency across the six open-source models. We consider two types of errors: \emph{single-emotion errors} (each of the 8 emotions, as a false negative or false positive: 16 candidates in total) and \emph{emotion-pair co-occurrence violations} (the 28 emotion pairs where both labels are present in the ground truth but the model predicts only one of them). The top two patterns from each category give us the four dominant error modes \textbf{P1}--\textbf{P4} (refer Table~\ref{tab:error_taxonomy} in Appendix). The counts in the table are aggregated across all six open-source models; the illustrative scenarios that follow are drawn from the best-performing baseline, Gemma-2-9B (Macro F1 = 0.500).

\paragraph{Single-emotion errors (P1, P2).}
The most common single-emotion failure is \emph{Trust under-prediction}. Every model misses Trust in at least 1{,}133 of the 4{,}731 scenarios, and this pattern holds across model families and sizes. This is consistent with the low inter-annotator agreement we observed for Trust (Fleiss $\kappa = 0.272$): Trust is at the same time the most prevalent emotion in EmoScene, the hardest emotion to label consistently, and the most-missed emotion by LLMs. The most common over-prediction is \emph{Anticipation}: models often read forward-looking language as Anticipation even when the underlying emotion is something else. The boxes P1 and P2 below show one representative example of each.

\begin{tcolorbox}[scenariobox, title=\textbf{P1: Trust under-prediction}]
\textit{True:} Trust. \quad \textit{Baseline:} (none).\\[3pt]
Keisha noticed the baby starting to cry in its stroller, and \textbf{without a moment's hesitation, she bent down with a caring gaze}. As she gently spoke soothing words, her brows furrowed slightly, scanning the small face for any signs of distress. The mother, looking flustered and juggling bags, caught Keisha's eye. \textbf{Keisha offered to hold the baby}, suggesting the mother take a quick break to collect herself, which she gratefully accepted.
\end{tcolorbox}

The model's own reasoning describes the actions accurately as ``caring and helpful towards the distressed baby and the flustered mother,'' but then concludes ``there's no direct indication that Keisha feels trust.'' The model recognises the caring behaviour, but treats Trust as something that needs to be explicitly named in the text. In Plutchik's framework, however, this kind of warm relational gesture -- offering help, attending to a stranger's distress -- is precisely what Trust looks like.

\begin{tcolorbox}[scenariobox, title=\textbf{P2: Anticipation over-prediction}]
\textit{True:} Fear. \quad \textit{Baseline:} Fear, Anticipation, Sadness.\\[3pt]
Sylvia sat quietly in the corner of the bustling caf\'e, stirring her coffee absentmindedly and \textbf{avoiding eye contact with the lively group nearby}. She watched them laugh and chat effortlessly, \textbf{feeling her palms dampen} as she considered \textbf{what to say if she joined them}. \textbf{Memories of a recent awkward exchange resurfaced}, making her glance away and fidget with the napkin in her lap. She knew she needed to practice more, so she decided to start small, smiling softly at the barista as she left her tip.
\end{tcolorbox}
The dominant emotion here is Fear -- social anxiety, indicated by the dampening palms, the avoided eye contact, and the memory of an awkward exchange. The model correctly picks up Fear, but it also predicts Anticipation, and the model's own reasoning makes the confusion explicit: it states that ``Sylvia is considering joining the group and feels anxious about it, which suggests she might be anticipating the interaction.'' But anxious imagining is not anticipation. Sylvia is not planning a future event; she is paralysed by social fear and turning the situation over in her head. The model conflates the conditional ``what to say if she joined them'' with eager planning. At the emotion-pair level, the top two co-occurrence violations (P3: Anticipation+Trust, P4: Joy+Trust) both involve Trust as the missed label. Representative examples and detailed analysis are provided in Appendix \ref{app:cooccurrence_errors}

\paragraph{Takeaway: the Trust convergence.}
In EmoScene, Trust is the most prevalent emotion ($48.1\%$ of scenarios) yet has the \emph{lowest} inter-annotator agreement (Fleiss $\kappa = 0.272$), reflecting genuine annotator disagreement on what counts as Trust. In Figure~\ref{fig:precision_recall_heatmap}, Trust shows the largest gap between its prevalence and its per-model recall. In the error analysis above, three of the four dominant error patterns (P1, P3, P4) are Trust-related. The implication is the same in all three places: Trust is signalled \emph{relationally and behaviourally} in human language -- offering help, listening attentively, easing posture -- but LLMs treat it as something that has to be explicitly named. The entanglement prior recovers Trust precisely where it co-occurs with cues the model \emph{does} catch (Joy, Anticipation), which is the mechanism behind BI's gains on Qwen2.5-7B and Gemma-2-2B.

\renewcommand{\tabularxcolumn}[1]{m{#1}}


\subsection{Linear-only Prior Ablation}
\label{app:linear_only_ablation}

The BI Ising prior combines a \emph{linear} (marginal) term $\sum_i \theta_i E_i$ and a \emph{pairwise} (entanglement) term $\sum_{i<j} \theta_{ij} E_i E_j$. To isolate the entanglement contribution, we ablate the pairwise dependencies ($\theta_{ij}=0$) at each model's F1-optimal $\alpha$ (Table~\ref{tab:linear_only_ablation}). For the models where BI is most effective, this pairwise term accounts for $32\text{--}34\%$ of the total absolute BI gain ($+1.7$ F1 for Qwen2.5-7B; $+1.2$ for Gemma-2-2B). Relative to the linear-only baseline, incorporating entanglement boosts F1 gains by \textbf{48--55\%} (from $+3.3$ to $+5.1$ for Qwen2.5-7B, and $+2.5$ to $+3.7$ for Gemma-2-2B). This confirms that while the marginal term drives the bulk of the improvement, the full Ising prior is critical for maximizing performance.




%% file: 07_summary_and_conclusion.tex
\section{Conclusion and Future Work}
\label{sec:concl}
We introduced \textbf{EmoScene}, a context-rich benchmark for evaluating multi-dimensional emotion understanding in language models. EmoScene represents emotional states using an 8-dimensional emotion vector and contains 4,731 scenarios that require reasoning over contextual and relational cues. Our evaluation of six instruction-tuned LLMs in chain-of-thought \& zero-shot setting shows that even strong models achieve modest performance, highlighting the difficulty of context-aware multi-label emotion prediction. To address the independence assumption commonly made in emotion classification, we proposed an entanglement-aware Bayesian inference framework that incorporates emotion co-occurrence statistics to perform joint posterior inference, improving prediction consistency and yielding notable gains for open-source models. These results suggest that modelling dependencies among emotions is important for robust emotion understanding. In future work, we plan to extend EmoScene with additional modalities and languages and investigate architectures and reasoning strategies that explicitly model emotional structure in context. Further, we would also like to explore the utility of preference tuning strategies in improving the emotional awareness of LLMs. 

%% file: 08_limitations.tex
\section*{Limitations}
\label{sec:limitations}
While EmoScene provides a context-rich benchmark for multi-dimensional emotion understanding, several limitations remain. First, the scenarios are generated using a language model and may not fully capture the diversity and complexity of naturally occurring emotional narratives. Second, the dataset focuses solely on textual descriptions and does not incorporate multimodal signals such as facial expressions, speech, or physiological cues that often accompany emotional expression. Finally, although our Bayesian inference framework models pairwise dependencies among emotions, it assumes relatively simple statistical relationships and may not capture higher-order emotional interactions present in real-world settings.
\section*{Ethics Statement}
\label{sec:ethics}
All snippets and examples used in this work were collected or generated following ethical practices and, where applicable, under permissive licenses. The dataset does not contain personally identifiable information or sensitive data. All human annotators involved in constructing the EmoScene dataset participated voluntarily and provided informed consent prior to annotation. Annotators were compensated on an hourly basis in accordance with the labor regulations of their respective country, and the task was designed to minimize cognitive burden during labeling. The scenarios were generated using a language model and subsequently reviewed and annotated by human experts, with duplicates and inconsistent outputs removed during dataset curation. Additionally, generative AI tools were used only to improve the clarity and grammar of the manuscript during writing; they were not used for literature search, research design, or idea generation. All scientific contributions, analyses, and conclusions presented in this work are the result of the authors’ own research.

%% file: 09_appendix.tex
\section{Dataset Statistics}
\label{app:dataset_statistics}
Table~\ref{tab: scene_stats} summarises the macro statistics of EmoScene. Scenarios contain rich contextual descriptions with an average length of \textbf{82 words} across roughly 4 sentences each, drawing on a vocabulary of \textbf{13{,}869 unique tokens}. About \textbf{60.9\%} of scenarios carry more than one emotion label, and 8.3\% are labelled with no emotion (the annotators judged that no Plutchik basic emotion was clearly conveyed). The mean label cardinality is \textbf{1.77}. Figure~\ref{fig:emoscene-label-count} shows the marginal label count per emotion.





\section{Dataset Analysis}
\label{app:dataset_analysis}
To ensure the robustness of the benchmark, we performed extensive post-hoc analyses on topic distribution and label skew. We verified thematic breadth by clustering scenarios. Additionally, we analyze the prevalence of Anticipation and Trust in our dataset, attributing their frequency to the inherent temporal and interpersonal nature of the source ECoK events.

\subsection{Topic diversity.}
\label{app:topic_diversity}

\begin{figure}[t]
\centering
\includegraphics[width=\columnwidth]{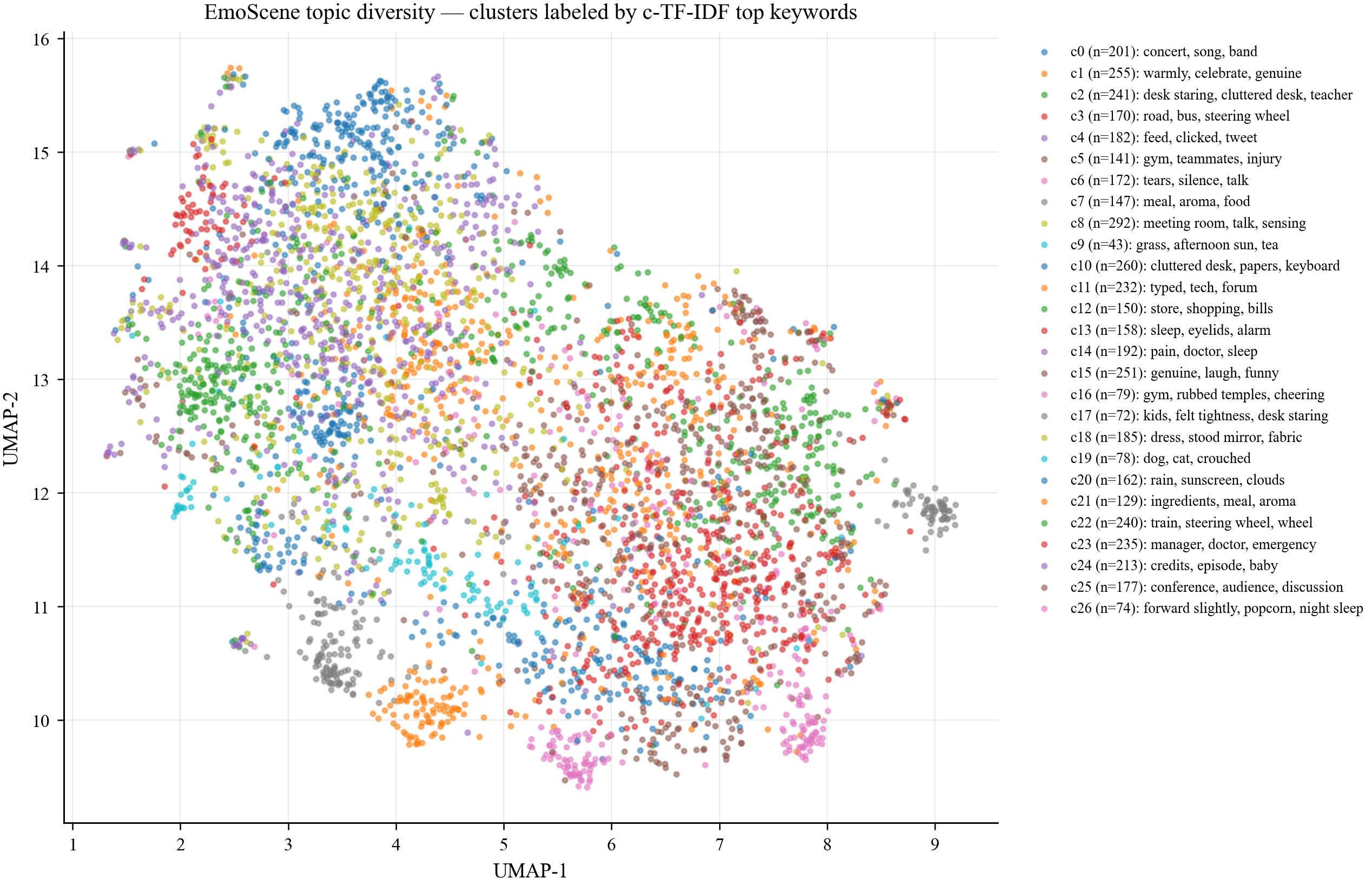}
\caption{UMAP projection of SBERT embeddings of all 4{,}731 EmoScene scenarios, coloured by KMeans cluster ($k=27$, Elbow optimized). Clusters separate cleanly into distinct topical regions.}
\label{fig:topic_umap}
\end{figure}

To verify that EmoScene covers a diverse range of everyday situations rather than a narrow topical band, we embed all 4{,}731 scenarios using SBERT framework \cite{reimers-2019-sentence-bert} with the \texttt{all-MiniLM-L6-v2} encoder and cluster their 384-dimensional embeddings into 27 groups via KMeans. The resulting clusters range in size from 43 to 292 scenarios and span clearly distinct themes: music and concerts, celebrations, school and study, driving and transport, social media, sports and injury, emotional talk, cooking and food, meetings and work, outdoors and nature, office and productivity, technology, errands and shopping, rest and fatigue, health and illness, humor and bonding, fitness, parenting and children, fashion and appearance, pets, weather, travel and commute, emergency and medical, entertainment and media, academic discourse, and nighttime leisure. We interpret these clusters deterministically by extracting class-conditional TF-IDF (c-TF-IDF) keywords, excluding common stop-words and proper nouns (tokens capitalised $>70\%$ of the time). The resulting cluster keywords and dominant emotions are detailed in Table~\ref{tab:topic_clusters}, with a 2-D UMAP projection of the embeddings shown in Figure~\ref{fig:topic_umap}.




\begin{figure}
    \centering
    \includegraphics[width=0.45\textwidth]{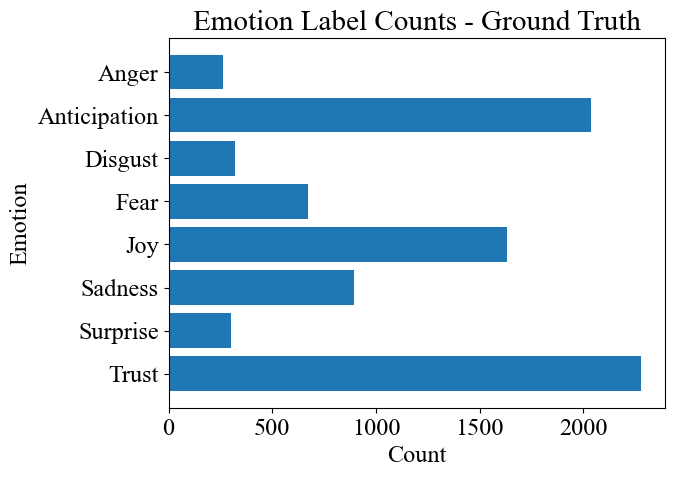}
    \caption{Emotion Label counts from the EmoScene Dataset.}
    \label{fig:emoscene-label-count}
\end{figure}

\subsection{Anticipation and Trust over-representation.}
\label{sec:anticipation_trust_over_representations}
Figure~\ref{fig:emoscene-label-count} for the label counts in EmoScene. Two emotions are over-represented in the marginal label distribution: Anticipation (43.1\%) and Trust (48.1\%). This reflects the structure of the underlying scenes rather than annotation drift. ECoK's source events are inherently temporal: each event ties together a cause, an outcome, and an action tendency, so scenarios derived from them naturally contain forward-looking narrative elements. Empirically, scenarios labelled Anticipation contain forward-looking lexical markers (\textit{soon, will, plan, hope, prepare,} etc.) at roughly \textbf{1.7$\times$} the rate of non-Anticipation scenarios. Trust over-representation similarly tracks ECoK's emphasis on bodily-symptom and feeling relations, which place scenarios in interpersonal contexts.

\begin{table}[t]
\centering
\small
\renewcommand{\arraystretch}{1.5}
\resizebox{\columnwidth}{!}{%
\begin{tabular}{@{}p{0.65\columnwidth}p{0.25\columnwidth}@{}}
\toprule
\textbf{Statistic} & \textbf{Value} \\
\midrule
Total number of scenarios       & 4,731 \\
Avg.\ scenario length (words)    & 82 \\
Median scenario length (words)  & 82 \\
Max scenario length (words)     & 129 \\
Min scenario length (words)     & 49 \\
Avg.\ sentences per scenario     & 4.24 \\
Vocabulary size (unique tokens) & 13,869 \\
Zero-emotion scenarios          & 395 (8.3\%) \\
Single-emotion scenarios        & 1,455 (30.7\%) \\
Multi-label scenarios ($\geq 2$) & 2,881 (60.9\%) \\
Average label cardinality       & 1.77 \\
\bottomrule
\end{tabular}%
}
\caption{Statistics of the EmoScene dataset.}
\label{tab: scene_stats}
\end{table}

\begin{table}[t]
\centering
\footnotesize
\setlength{\tabcolsep}{4pt}
\begin{tabular}{l r p{0.42\columnwidth}}
\toprule
\textbf{Cluster} & \textbf{Size} & \textbf{Top c-TF-IDF keywords} \\
\midrule
0 -- Music / concerts       & 201 & concert, song, band \\
1 -- Celebrations           & 255 & warmly, celebrate, genuine \\
2 -- School / study         & 241 & desk staring, cluttered desk, teacher \\
3 -- Driving / transport    & 170 & road, bus, steering wheel \\
4 -- Social media           & 182 & feed, clicked, tweet \\
5 -- Sports / injury        & 141 & gym, teammates, injury \\
6 -- Emotional talk         & 172 & tears, silence, talk \\
7 -- Cooking / food         & 147 & meal, aroma, food \\
8 -- Meetings / work        & 292 & meeting room, talk, sensing \\
9 -- Outdoors / nature      & 43  & grass, afternoon sun, tea \\
10 -- Office / productivity & 260 & cluttered desk, papers, keyboard \\
11 -- Technology            & 232 & typed, tech, forum \\
12 -- Errands / shopping    & 150 & store, shopping, bills \\
13 -- Rest / fatigue        & 158 & sleep, eyelids, alarm \\
14 -- Health / illness      & 192 & pain, doctor, sleep \\
15 -- Humor / bonding       & 251 & genuine, laugh, funny \\
16 -- Fitness               & 79  & gym, rubbed temples, cheering \\
17 -- Parenting / children  & 72  & kids, felt tightness, desk staring \\
18 -- Fashion / appearance  & 185 & dress, stood mirror, fabric \\
19 -- Pets                  & 78  & dog, cat, crouched \\
20 -- Weather / outdoors    & 162 & rain, sunscreen, clouds \\
21 -- Cooking (detail)      & 129 & ingredients, meal, aroma \\
22 -- Travel / commute      & 240 & train, steering wheel, wheel \\
23 -- Emergency / medical   & 235 & manager, doctor, emergency \\
24 -- Entertainment / media & 213 & credits, episode, baby \\
25 -- Academic / discourse  & 177 & conference, audience, discussion \\
26 -- Nighttime / leisure   & 74  & forward slightly, popcorn, night sleep \\
\bottomrule
\end{tabular}
\caption{Topic clusters (k=27, elbow-optimised) of EmoScene scenarios derived from KMeans on SBERT embeddings, labelled by c-TF-IDF top keywords. The high optimal k confirms fine-grained topical diversity across the corpus.}
\label{tab:topic_clusters}
\end{table}

\section{EmoScene Emotion Entanglement}
    
\includegraphics[width=0.45\textwidth]{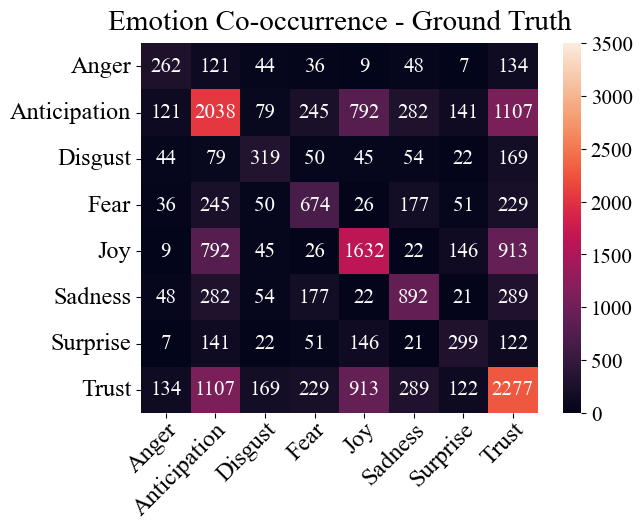}

Above, we see the co-occurrence of emotion labels in EmoScene dataset. The frequency of two given emotions being assigned to the same Scenario. It appears that anticipation, joy, and trust co-occur quite frequently, meaning our models may confuse between them. Their co-occurrences might also account for their high individual counts in our dataset overall.

\subsection{Empirical Validation of Entanglement Prior against Plutchik's Theory}
\label{app:plutchik_alignment}

\begin{figure}[t]
\centering
\includegraphics[width=0.95\linewidth]{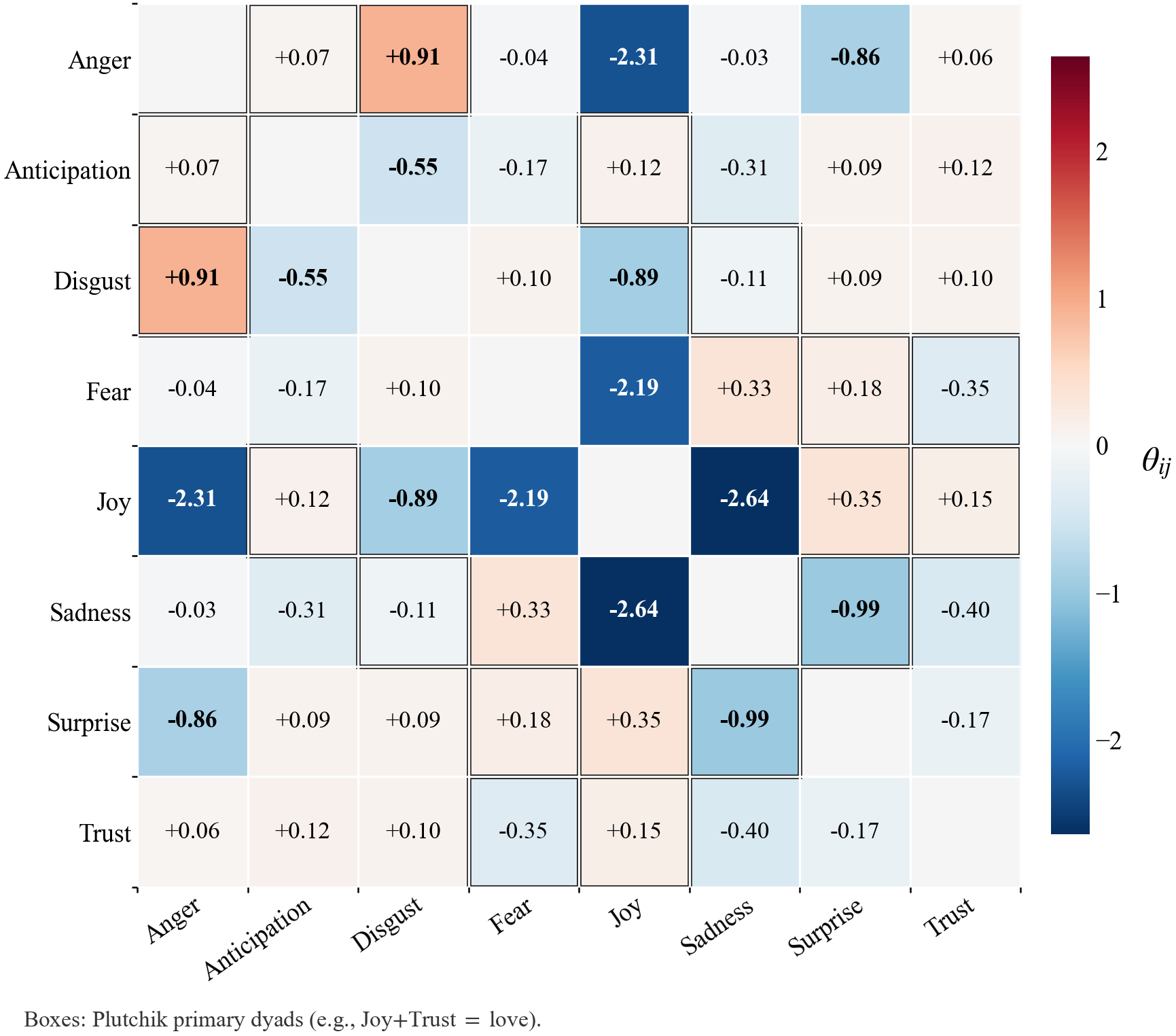}
\caption{Learned pairwise couplings $\theta_{ij}$ on EmoScene. Red cells indicate positive couplings (the pair co-occurs more often than under independence) and blue cells indicate negative couplings (mutual exclusion). The eight cells outlined in black correspond to Plutchik's primary dyads (adjacent on his wheel; e.g., Joy+Trust = ``love''). Five of the eight primary dyads carry positive couplings, including the strongest overall (Disgust+Anger = ``contempt'', $+0.91$), and the largest negative coupling (Joy+Sadness, $-2.64$) coincides with Plutchik's primary bipolar axis. The remaining three primary dyads (``submission'', ``disapproval'', ``remorse'') appear with negative sign, reflecting corpus-specific deviations from Plutchik's symmetric prediction.}

\label{fig:theta_ij_heatmap}
\end{figure}

Figure~\ref{fig:theta_ij_heatmap} visualises the learned $\theta_{ij}$ matrix. The structure of the prior is not an arbitrary statistical artifact of EmoScene -- it operationalises Plutchik's wheel. The strongest positive couplings are exactly the pairs Plutchik identified as primary dyads: \emph{Anger + Disgust} ($\theta_{ij} = +0.91$, his ``contempt'' dyad), \emph{Joy + Trust} ($+0.15$, the ``love'' dyad), \emph{Anticipation + Trust} ($+0.12$), and \emph{Fear + Sadness} ($+0.33$). Symmetrically, the strongest negative couplings fall on Plutchik's opposite axes: \emph{Joy $\leftrightarrow$ Sadness} ($-2.64$), \emph{Anger $\leftrightarrow$ Joy} ($-2.31$), \emph{Fear $\leftrightarrow$ Joy} ($-2.19$). The prior is therefore Plutchik-consistent by data, not by construction: the wheel structure emerges directly from EmoScene's annotation co-occurrence. This is the empirical link between our theoretical positioning (\S\ref{sec:03_emoscene_dataset}) and the inference machinery -- the prior $\theta_{ij}$ is precisely what makes the Bayesian step ``entanglement-aware'' in Plutchik's sense, and an ablation removing it loses 48--52\% of BI's F1 gain on the models where BI is most effective (Appendix~\ref{app:linear_only_ablation}).

\section{Extended Experimental setup}
\label{app:extended_exp_setup}
This section contains the details about Prompting Strategies that are used in the experiments.

\paragraph{Prompting and Decoding.} We evaluate two configurations: a \textbf{chain-of-thought (CoT)} prompt that elicits step-by-step reasoning and a self-reported confidence score (1--5) before outputting a yes/no label, and a \textbf{zero-shot (ZS)} prompt that requires only the label. For open-source models, we use greedy decoding ($T=0$) with a max limit of 512 tokens for CoT and 16 tokens for ZS.

\section{Extended Results \& Analysis}
\label{app:extended_results}
This section includes details about the effect of $\alpha$ and Sensitivity of prompting strategies.

\subsection{Per-Emotion Performance}
\label{sec:per_emotion}
\begin{figure*}[t]
    \centering
    \begin{subfigure}{0.48\textwidth}
        \centering
        \includegraphics[width=\linewidth]{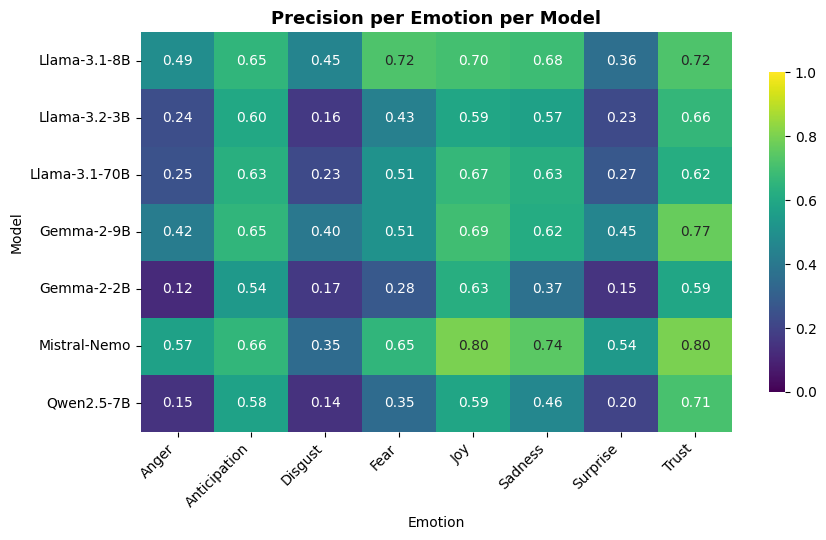}
        \caption{Precision per Emotion per Model}
        \label{fig:precision_heatmap}
    \end{subfigure}
    \hfill 
    \begin{subfigure}{0.48\textwidth}
        \centering
        \includegraphics[width=\linewidth]{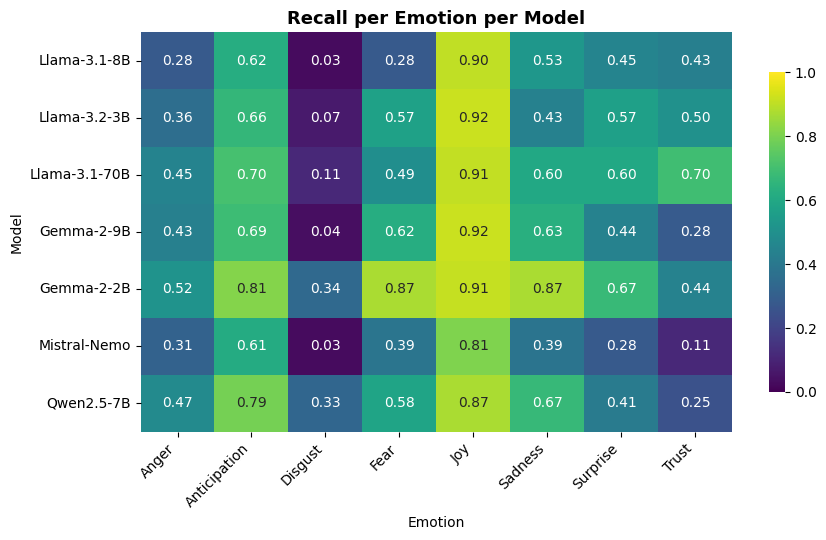}
        \caption{Recall per Emotion per Model}
        \label{fig:recall_heatmap}
    \end{subfigure}
    
    \caption{Heatmap comparison of Precision and Recall for Plutchik's 8 basic emotions across all evaluated models. Darker regions indicate lower performance, while lighter/greener regions indicate higher alignment.}
    \label{fig:precision_recall_heatmap}
\end{figure*}

Per-emotion precision and recall heatmaps for all seven open-source models are shown in Figure~\ref{fig:precision_recall_heatmap}. Three patterns are consistent across the model panel. \textbf{Joy} achieves high precision and recall across every model; lexical cues for joy (smile, laughter, warmth) are reliable and frequent in EmoScene. \textbf{Anticipation} shows moderate precision and recall: it is over-predicted by smaller models (Gemma-2-2B), consistent with the false-positive pattern P2 in \S\ref{sec:error_analysis}. \textbf{Disgust} and \textbf{Surprise} suffer critically low recall (near-zero in Llama-3.2-3B and Mistral-Nemo), indicating models predict these emotions only when evidence is  (abundant). This is the over-prediction--vs--conservatism axis that BI exploits: the marginal prior $\theta_i$ pulls baselines that over-predict rare emotions (Qwen2.5-7B, Gemma-2-2B) toward the empirical zero-rate, raising precision without losing the high-recall Joy and Anticipation cells.

\subsection{Significance of \texorpdfstring{$\alpha$}{alpha}}
\label{sec:alpha_sensitivity}

\begin{figure*}[t]
\centering
\includegraphics[width=0.95\textwidth]{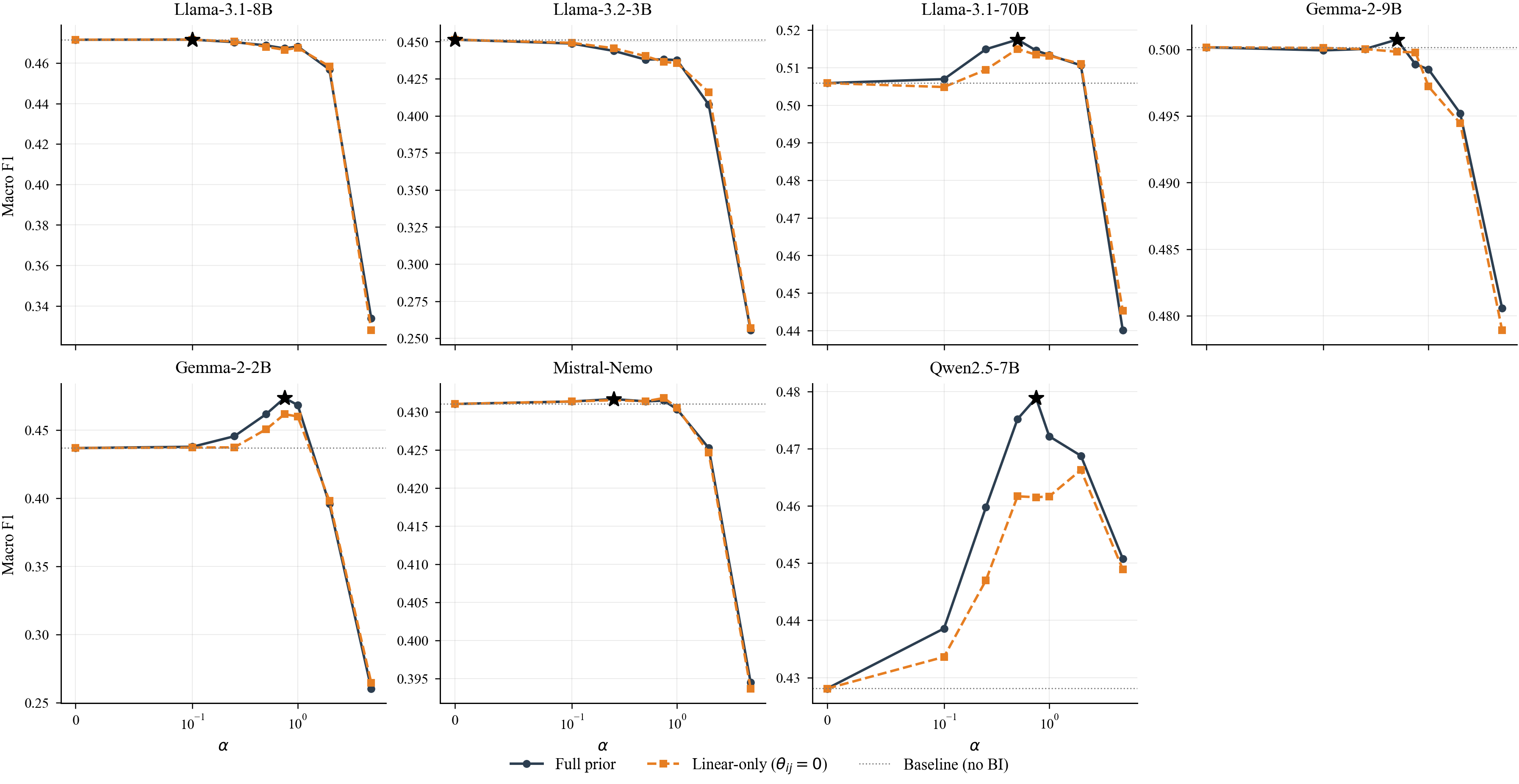}
\caption{Macro F1 as a function of $\alpha$ for each open-source model under two prior configurations: the full entanglement-aware prior (solid) and the linear-only ablation with $\theta_{ij}=0$ (dashed). Black stars mark the F1-optimal $\alpha$ used in Table~\ref{tab:complete_results_table}; the dotted horizontal line marks the baseline ($\alpha=0$).}
\label{fig:alpha_sensitivity}
\end{figure*}

The hyperparameter $\alpha$ in our MAP objective controls how strongly the entanglement prior pulls the model's marginal predictions toward the empirical co-occurrence structure of EmoScene. We sweep $\alpha$ over $\{0, 0.1, 0.25, 0.5, 0.75, 1, 2, 5\}$ for each model and report the per-model F1-optimal value (the same value is used for Table~\ref{tab:complete_results_table}). Figure~\ref{fig:alpha_sensitivity} plots Macro F1 as a function of $\alpha$ for each model, alongside a linear-only ablation in which the pairwise term $\theta_{ij}$ is held at zero.

Three observations follow. \textbf{(1) Smooth curves.} The F1-vs-$\alpha$ curve is smooth and well-behaved for every model, with no narrow peak. BI's gains are not the artefact of fine-tuning a single value of $\alpha$; small mis-calibrations in either direction leave the result essentially unchanged. \textbf{(2) Llama-3.2-3B's flat curve explains its zero improvement.} For Llama-3.2-3B, Macro F1 stays at the baseline value across all $\alpha$, so the F1-optimal $\alpha = 0$ is a benign symptom of an inert response rather than a sign of $\alpha$-shopping or BI failing to engage. This model's marginal predictions are already well-calibrated to EmoScene's emotion distribution, so the prior $\theta_{ij}$ has nothing to add. \textbf{(3) The pairwise term contributes meaningfully where BI helps.} For the two over-predicting baselines (Qwen2.5-7B, Gemma-2-2B), the dashed linear-only curve sits visibly below the solid full-prior curve at the operating $\alpha$; an ablation (Appendix \ref{app:linear_only_ablation}) quantifies that removing $\theta_{ij}$ at the operating $\alpha$ erases 48--52\% of BI's F1 gain. The entanglement term is therefore not decorative -- on the models where BI is effective, roughly half of the gain comes specifically from modelling pairwise emotion co-occurrence.

\subsection{Effect of Prompting Strategy}
\label{sec:prompting_strategy}
The results in Table~\ref{tab:complete_results_table} contains results corresponding to two prompt variants. First, chain-of-thought that asks for step-by-step reasoning along with the answer and other is the zero-shot that only asks for answer without any intermediate reasoning.

Two observations follow. First, BI provides positive aggregate gains under \emph{both} prompting strategies, indicating that the benefit of modeling emotion entanglement is not specific to any particular reasoning style. Second, several models attain a higher baseline F1 under the no-CoT prompt (e.g., Llama-3.1-8B: 0.505 vs.\ 0.472; Gemma-2-9b: 0.506 vs.\ 0.500), suggesting that step-by-step reasoning is not uniformly helpful for this task and can amplify systematic biases in the model's emotion priors.

\begin{table*}[t]
\centering
\footnotesize
\setlength{\tabcolsep}{3pt}
\resizebox{\textwidth}{!}{%
\begin{tabular}{l rrrrrrr}
\toprule
\textbf{Pattern} & \textbf{Llama-3.1-8B} & \textbf{Llama-3.2-3B} & \textbf{Llama-3.1-70B} & \textbf{Gemma-2-9B} & \textbf{Gemma-2-2B} & \textbf{Mistral-Nemo} & \textbf{Qwen2.5-7B} \\
\midrule
P1: Trust under-prediction (FN)               & 1307 & 1133 & 694 & 1629 & 1270 & 2020 & 1699 \\
P2: Anticipation over-prediction (FP)         &  677 &  904 & 842 & 738 & 1424 &  650 & 1190 \\
P3: Anticipation\,+\,Trust co-occurrence miss &  557 &  500 & 428 & 644 &  596 &  647 &  734 \\
P4: Joy\,+\,Trust co-occurrence miss          &  392 &  269 & 263 & 529 &  415 &  640 &  586 \\
\bottomrule
\end{tabular}%
}
\caption{Incidence of the four dominant error patterns across six baseline LLMs on EmoScene. Each pattern is defined by a deterministic Boolean rule on the prediction array; counts are over all 4{,}731 scenarios. \textbf{P1}--\textbf{P2} are the top-2 single-emotion error modes (ranked across 8 emotions $\times$ \{FN, FP\}); \textbf{P3}--\textbf{P4} are the top-2 emotion-pair co-occurrence violations (ranked across the 28 emotion pairs).}
\label{tab:error_taxonomy}
\end{table*}

\subsection{Confidence and Calibration}
\label{sec:calibration}

Under our CoT prompt each model produces two confidence signals: a self-reported integer in $\{1,2,3,4,5\}$ inside the $\langle$confidence$\rangle$ tags, and a logit-derived value $\max(P_{\text{yes}}, P_{\text{no}})$ computed from the softmax at the answer token. We measure how well each signal tracks actual accuracy on EmoScene using the Expected Calibration Error (ECE). Per-model ECE values are reported in Appendix~\ref{app:calibration}.

At the token level, all six open-source models are over-confident. Mean logit-derived confidence ranges from $0.92$ to $0.998$ across models, while their actual accuracy is only $0.71$ to $0.84$. Token-level ECE ranges from $0.149$ (Llama-3.1-8B) to $0.232$ (Gemma-2-2B). Self-reported confidence shows a different pattern: some models are under-confident (Llama-3.1-8B reports a mean of $0.47$/$1.0$ but is accurate $0.84$ of the time; Qwen2.5-7B reports $0.43$ and is accurate $0.76$), while Mistral-Nemo is nearly perfectly calibrated. 

The most useful finding for BI is in Table~\ref{tab:bi_by_confidence}. For the two models where BI is effective, the gain scales directly with baseline uncertainty. In the lowest-confidence quartile (Q1), BI corrects an additional $1.06$ (Qwen2.5-7B) to $1.49$ (Gemma-2-2B) emotion cells per scenario on average. In the highest-confidence quartile (Q4), the same gain shrinks to $0.03$ and $0.05$. The decrease is monotonic across all four quartiles. This trend is expected is consistent with the behaviour from an uncertainity-aware prior: \textbf{BI provides the largest improvements in regions where the baseline model exhibits lower confidence}. For models whose baseline is uniformly high-confidence (Gemma-2-9B, Llama-3.1-8B), the same per-quartile gains are essentially zero -- there is no uncertainty for the prior to compensate, which matches their small overall BI gains in Table~\ref{tab:complete_results_table}.

\begin{table}[t]
\centering
\footnotesize
\setlength{\tabcolsep}{3pt}
\begin{tabular}{l c c c c}
\toprule
\textbf{Model} & \textbf{Q1 (lowest)} & \textbf{Q2} & \textbf{Q3} & \textbf{Q4 (highest)} \\
\midrule
Qwen2.5-7B    & $+1.06$ & $+0.60$ & $+0.31$ & $+0.03$ \\
Gemma-2-2B    & $+1.49$ & $+0.64$ & $+0.33$ & $+0.05$ \\
\midrule
Gemma-2-9B    & $+0.04$ & $\sim 0$ & $\sim 0$ & $\sim 0$ \\
Llama-3.1-8B  & $\sim 0$ & $\sim 0$ & $\sim 0$ & $\sim 0$ \\
\bottomrule
\end{tabular}
\caption{Mean per-scenario BI gain ($\Delta$correct-emotion-cells, out of 8) within each quartile of the baseline's logit-derived confidence. For the two models where BI is most effective (top), the gain decreases monotonically with confidence. For uniformly high-confidence baselines (bottom), the gain is small in every quartile.}
\label{tab:bi_by_confidence}
\end{table}

\subsection{Co-occurrence Error Patterns}
\label{app:cooccurrence_errors}
At the emotion-pair level, the top two failure modes both involve Trust. Anticipation+Trust (P3) and Joy+Trust (P4) are the two pairs most often broken by the baseline, where the model predicts one of the pair but not the other. Together, P3 and P4 covers 661 to 1{,}320 scenarios per model, which is comparable in scale to the single-emotion patterns. These are exactly the failure modes that our entanglement term $\theta_{ij}$ is designed to fix. As we show in the Appendix, removing $\theta_{ij}$ removes 48--52\% of BI's F1 gain on the models where BI is most effective. The boxes P3 and P4 below show one example of each.

\begin{tcolorbox}[scenariobox, title=\textbf{P3: Anticipation\,+\,Trust co-occurrence miss}]
\textit{True:} Anticipation, Trust. \quad \textit{Baseline:} Anticipation.\\[3pt]
Edna stood on the edge of the field, part of a group of fellow football players, her new gear snug and ready for \textbf{the upcoming match}. Her heart was pounding steadily, and a light sheen of sweat glistened on her brow. She took a deep breath, muscles taut, and began her stretching routine, glancing occasionally towards her teammates. Edna's jaw was clenched as \textbf{she listened to their captain strategize, absorbing every word, her eyes steady and sure}.
\end{tcolorbox}
The model correctly picks up Anticipation (``her physical symptoms and her focused attention on the upcoming match and her captain's strategy all point towards anticipation''), but it misses Trust. Its own reasoning explains why: ``these actions could indicate respect or determination, [but] there's no direct evidence suggesting trust towards her teammates or the captain.'' The captain-team listening dynamic -- absorbing every word, a steady gaze -- is exactly the kind of relational Trust the entanglement prior $\theta_{ij}$ is designed to recover. In the learned prior, Anticipation and Trust are positively coupled ($\theta_{ij} = +0.12$): whenever Anticipation is correctly identified in a team setting, the coupling raises the posterior on Trust as well.

\begin{tcolorbox}[scenariobox, title=\textbf{P4: Joy\,+\,Trust co-occurrence miss}]
\textit{True:} Joy, Trust. \quad \textit{Baseline:} Joy.\\[3pt]
John stood with Anna in the bustling caf\'e, their eyes meeting for the first time in weeks. \textbf{Anna smiled warmly and remarked on John's striking new jacket}, noting how well it suited him. As she spoke, \textbf{John's face felt warmer, a smile spreading effortlessly across his lips, his posture easing}. He nodded in thanks, deciding he would keep experimenting with different styles, maybe even following a few fashion influencers for new ideas.
\end{tcolorbox}
The model correctly identifies Joy from the warm smile and easing posture, but misses Trust with the reasoning that ``while warmth and positive feelings are present, there's no explicit indication of trust being established or felt.'' Yet John's posture easing in Anna's presence -- and his open willingness to take her suggestion -- is precisely the kind of physical comfort that signals Trust in an intimate exchange. Joy and Trust together form Plutchik's ``love'' dyad, and the entanglement prior $\theta_{ij}$ is positively coupled here ($\theta_{ij} = +0.15$): whenever Joy is identified in a warm interpersonal exchange, the prior pulls the posterior on Trust upward, recovering exactly this kind of miss.

\section{Confidence Analysis}
\label{app:calibration}

We measure calibration of each model's confidence signals against actual accuracy on EmoScene using the Expected Calibration Error (ECE). Two confidence signals are available under the CoT prompt: a self-reported integer in $\{1,2,3,4,5\}$ from the $\langle$confidence$\rangle$ tags, and a logit-derived value $\max(P_{\text{yes}}, P_{\text{no}})$ from the softmax at the answer token. Per-model ECE values are listed in Table~\ref{tab:ece_summary}.

\begin{table}[t]
\centering
\footnotesize
\setlength{\tabcolsep}{5pt}
\begin{tabular}{l c c}
\toprule
\textbf{Model} & \textbf{Self-reported ECE} & \textbf{Logit-derived ECE} \\
\midrule
Llama-3.1-8B          & 0.400 & 0.149 \\
Llama-3.2-3B          & 0.158 & 0.175 \\
Llama-3.1-70B         & 0.075 & 0.145 \\
Mistral-Nemo          & 0.090 & 0.164 \\
Qwen2.5-7B            & 0.370 & 0.163 \\
Gemma-2-2B            & 0.118 & 0.232 \\
Gemma-2-9B            & 0.306 & 0.162 \\
\bottomrule
\end{tabular}
\caption{Expected Calibration Error for each model under the CoT prompt.}
\label{tab:ece_summary}
\end{table}

At the token level (logit-derived), every open-source model is \emph{over-confident}: mean confidence ranges from $0.92$ to $0.998$, but actual accuracy is only $0.71$ to $0.84$. The token-level ECEs sit between $0.149$ (Llama-3.1-8B) and $0.232$ (Gemma-2-2B), with the gap concentrated in the highest-confidence bins -- the model says it is sure roughly $97$\% of the time but is wrong $15$--$30$\% of those cases. This is consistent across all six models at the logit level, current open-source LLMs are uniformly overconfident on subjective emotion judgements on EmoScene.

Self-reported confidence (the $\langle$confidence$\rangle$ tag) shows the opposite pattern for several models. Llama-3.1-8B reports a mean confidence of $0.47$/$1.0$ while being accurate $0.84$ of the time, and Qwen2.5-7B reports $0.43$ while being accurate $0.76$ -- both are noticeably \emph{under}-confident in their verbal reports relative to their actual accuracy (ECE = $0.40$ and $0.37$ respectively). Mistral-Nemo is nearly perfectly calibrated by self-report (ECE $0.090$ and $0.078$). The gap between the two signals -- token-level over-confidence and report-level under-confidence on the same model -- suggests that the verbal report is not a faithful reflection of the underlying token-level distribution. Our BI implementation uses the logit-level signal directly via the softmaxed Yes/No probabilities, sidestepping this disagreement.

\section{Scenario Generation Prompt}
\label{app:emoscene-generation-prompt}
The exact system and user prompts used to instruct the language model for synthetic scenario generation are detailed in Figure~\ref{fig:generation_prompt}. We have prompted GPT-4o to generate the scenarios. We have used a temperature of $0.9$, top\_p=$1.0$ \& max-tokens=$200$.

The evaluation prompt that we have used to evaluate the language models are provided in the Figure \ref{fig:evaluation_prompt}. 

\begin{figure*}[ht]
    \begin{tcolorbox}[colback=gray!5!white, colframe=black, title=\textbf{Scenario Generation Prompt}]
    \footnotesize
    
    \section*{System Message}
    You are an expert at transforming structured knowledge graph information into vivid, emotionally resonant, and realistic scenarios. A knowledge graph is represented as (head, relation, tail) triplets. Events describing emotions are regarded as head entities, while frame relations constitute edge‐type relations, with various corresponding attributes serving as tails. Note that a single event (head) can have multiple relations and tails. These relations are often interdependent, and the tails should be selected carefully to maintain narrative coherence and realism. You do not need to use every relation or tail—choose the ones that best support a believable and naturalistic scenario.

You will be given an event along with relations and their corresponding tails. Your task is to generate a concise, vivid, and realistic scenario (4-5 sentences MAXIMUM) that logically connects the head, relation, and tail. Avoid stylized, poetic, or abstract descriptions. **Do not use any feeling‐words from the “Feelings” relation directly in your sentences**—the emotion should be implied through actions, tone, or imagery. You may add at most one extra character if needed.

Procedure for Scenario Generation:
\begin{enumerate}
    \item Examine all available (relation, tail) pairs for the event.
    \item Identify and group correlated or complementary relation-tail pairs
    \item Enumerate different meaningful subsets of relation-tail pairs that can lead to a coherent narrative.
    \item Choose the subset that logically connects to the event and allows for a realistic and emotionally suggestive scenario.
    \item Construct a 4–5 sentence scenario using only the selected subset. Ensure the scenario has:
    \begin{itemize}
        \item A clear flow of events
        \item A realistic and natural naritive
        \item Strong coreference resolution
        \item Emotion shown through physical, verbal, or contextual cues
    \end{itemize}
\end{enumerate}

Rules:
\begin{itemize}
    \item Use **only** the provided relations/tails—*do not* invent new details.
    \item Write a coherent, vivid, and emotionally expressive scenario in 4–5 sentences.
    \item Use **simple, everyday English**. Avoid complex sentence structures, rare words, or overly formal or literary tone. The scenario should sound like something a regular person would say or observe.
    \item Do not use metaphors, figurative expressions, or abstract concepts—be literal and grounded.
    \item You must not use any word or phrase from the “Feelings” relation directly—convey the emotion through actions, tone, and imagery instead.
    \item Pronouns must refer clearly to the correct entity; maintain good coreference resolution.
\end{itemize}
    
    \section*{User Message}
    Event: \texttt{head}
    
    Do NOT use the following words in any form. These are emotion words from the Feelings relation: \texttt{feelings\_list}.
    
    Relations and Tails:
    \texttt{rel\_tail\_string}
    
    Use only clear, simple, and natural English. Avoid complex phrases.
    
    A concise, realistic scenario (4–5 sentences) using the most relevant relation-tail pairs. Emotion should be shown, not named.
    
    Begin:
    \end{tcolorbox}
    \caption{Scenario generation prompt used to instruct the language model.}
    \label{fig:generation_prompt}
\end{figure*}

\begin{figure*}[ht]
    \begin{tcolorbox}[colback=gray!5!white, colframe=black, title=\textbf{EmoScene Evaluation Prompt}]
    \footnotesize
    
    \section*{System Message}
    You are an expert in emotional analysis and natural language processing. Your task is to answer whether the subject might feel the particular emotion with a yes or no. Yes indicates that the subject experiences the emotion, while no indicates that the subject doesn't experience the emotion. Think step by step to arrive at the final answer. \\
        
        In your response, you must follow this exact sequence:
        \begin{enumerate}
        \item Provide a brief, step-by-step reasoning for your analysis.
        \item State your confidence level in your prediction on a scale of 1 to 5 using the rubric below, enclosed within <confidence></confidence> tags. DO NOT provide an explanation for your confidence level, just the number.
        \item Provide your final prediction (yes or no) between <answer></answer> tags. A "yes" indicates that the subject experiences the emotion, while a "no" indicates that the subject doesn't experience the emotion.
        \end{enumerate}

        Confidence Rubric:
        \begin{itemize}
            \item <confidence>1</confidence> : The scenario gives almost no clues about the emotion; it is highly ambiguous.
            \item <confidence>2</confidence> : The scenario gives weak or unclear hints that could easily indicate another emotion.
        \item <confidence>3</confidence> : The scenario contains typical conversational cues for the emotion but no strong evidence.
        \item <confidence>4</confidence> : Most elements in the scenario clearly suggest this emotion, with little ambiguity.
        \item <confidence>5</confidence> : The emotion is explicitly stated or the evidence is completely clear.
        \end{itemize}
    
    \section*{User Message}
    \textbf{Scenario}:\\
{text}

\textbf{Subject}: \\
{subject}

\textbf{Question}: \\
Does {subject} feel {emotion}?

\textbf{Note}: \\
Based on the above-provided information, answer the question. Please note that your output must follow the exact sequence outlined in the instructions: provide your brief reasoning, your 1-5 confidence score wrapped in <confidence></confidence> tags, and your final yes/no answer wrapped in <answer></answer> tags. DO NOT provide any explanation for your confidence score.
    \end{tcolorbox}
    \caption{EmoScene Evaluation Prompt.}
    \label{fig:evaluation_prompt}
\end{figure*}

\begin{figure*}[ht]
    \begin{tcolorbox}[colback=gray!5!white, colframe=black, title=\textbf{Zero-Shot Evaluation Prompt}]
    \footnotesize
    
    \section*{System Message}
    You are an expert in emotional analysis and natural language processing. Your task is to answer whether the subject might feel the particular emotion with a yes or no. Yes indicates that the subject experiences the emotion, while no indicates that the subject doesn't experience the emotion. \\
    
    You must answer strictly by wrapping your Yes or No prediction inside \texttt{<answer></answer>} tags. \\
    
    Do not provide any reasoning, explanation, or extra text. Output ONLY the tags and your answer.
    
    \section*{User Message}
    \textbf{Scenario}:\\
{text}

\textbf{Subject}: \\
{subject}

\textbf{Question}: \\
Does {subject} feel {emotion}?

\textbf{Note}: \\
Based on the above-provided information, answer the question strictly with Yes or No wrapped inside \texttt{<answer></answer>} tags. Do not provide any explanation, reasoning, or additional text.
    
    \end{tcolorbox}
    \caption{Zero-Shot Evaluation Prompt.}
    \label{fig:zeroshot_evaluation_prompt}
\end{figure*}
